\documentclass[conference]{IEEEtran}
\IEEEoverridecommandlockouts
\usepackage{cite}
\usepackage{amsmath,amssymb,amsfonts}
\usepackage{algorithmic}
\usepackage{graphicx}
\usepackage{textcomp}
\usepackage{xcolor}
\usepackage{booktabs}
\usepackage{makecell}
\usepackage{multirow}
\usepackage[colorlinks,linkcolor=blue]{hyperref}

\begin{document}

\title{A Survey of Video-based Action Quality Assessment \\
\thanks{*Corresponding author.

This work was supported by Shanghai Municipal Science and Technology Major Project 2021SHZDZX0103 and National Natural Science Foundation of China under Grant 82090052.}
}

\author{
\IEEEauthorblockN{
Shunli Wang\textsuperscript{1}, 
Dingkang Yang\textsuperscript{1},
Peng Zhai\textsuperscript{1}, 
Qing Yu\textsuperscript{2}, 
Tao Suo\textsuperscript{2},
Zhan Sun\textsuperscript{2},
Ka Li\textsuperscript{2},
Lihua Zhang\textsuperscript{3,1}*
}
\IEEEauthorblockA{
\textit{Institute of AI \& Robotics, Fudan University} \textsuperscript{1}, Shanghai, China  \\
\textit{ZhongShan Hospital} \textsuperscript{2}, Shanghai, China 
\ \ \ \textit{Ji Hua Laboratory} \textsuperscript{3}, Foshan, China \\
\{slwang19, dkyang20, pzhai18, lihuazhang\}@fudan.edu.cn \\\{ yu.qing, suo.tao, sun.zhan, li.ka \}@zs-hospital.sh.cn
}
}


\maketitle

\begin{abstract}
Human action recognition and analysis have great demand and important application significance in video surveillance, video retrieval, and human-computer interaction.
The task of human action quality evaluation requires the intelligent system to automatically and objectively evaluate the action completed by the human. 
The action quality assessment model can reduce the human and material resources spent in action evaluation and reduce subjectivity.
In this paper, we provide a comprehensive survey of existing papers on video-based action quality assessment.
Different from human action recognition, the application scenario of action quality assessment is relatively narrow. Most of the existing work focuses on sports and medical care. 
We first introduce the definition and challenges of human action quality assessment.
Then we present the existing datasets and evaluation metrics.
In addition, we summarized the methods of sports and medical care according to the model categories and publishing institutions according to the characteristics of the two fields.
At the end, combined with recent work, the promising development direction in action quality assessment is discussed.

\end{abstract}


\section{Introduction}\label{ch1}
With the prosperity and development of video processing technology and deep neural networks (DNNs), many studies have focused on human behaviour modelling, especially in human action recognition and analysis.
Existing models\cite{TwoStream,TwoStreamV2,I3D,C3D} have been able to complete complex tasks such as action classification of short videos, temporal action segmentation, classification of long-term videos and spatial-temporal action location.
These studies have also been widely used in video retrieval, anomaly detection, and entertainment.
However, these works can only classify and locate human action at a coarse granularity level and can not objectively evaluate the quality and compliance of some specific actions.

Based on the research of human action recognition, some studies have begun to pay attention to the action quality assessment (AQA) model.
The AQA task aims to design a system that can automatically and objectively evaluate some specific actions completed by people through input videos.
AQA has many practical application scenarios, such as surgical skill rating and medical rehabilitation test in medical care, athlete posture correction and coaching system in sports, operation compliance analysis and dangerous action monitoring in industrial production.
High-performance AQA systems can significantly improve the professional level of people, enhance training efficiency and reduce training costs.
Models in HAR tasks only need to capture the external differences between various action categories, while models in AQA tasks focus on distinguishing internal differences within a specific action. 
This characteristic makes the AQA task more complex than the HAR task and puts forward higher requirements for the perception ability of the model.

Nowadays, the research on AQA tasks is still in its infancy. 
The action recognition task can be easily extended to rich daily behaviors, while the AQA task only applies to specific professional actions. Therefore, it is easy to find that AQA has fewer application scenarios than action recognition.
Considering the data accessibility and task complexity, most studies focus on the action quality evaluation in sports and medical care. 
AQA methods discussed in this paper are also in these two fields.
Noted that due to space and topic limitation, we only focus on the video-based action assessment methods in this paper. Methods based on kinematic sensors are excluded.

This paper provides a systematic review of the progress of video-based human action quality assessment tasks for the first time. 
There are three major contributions of this paper:
\begin{itemize}
 \item The existing AQA work is comprehensively reviewed and summarized, including the critical work and progress in the field of AQA.
 \item This paper focuses on reviewing AQA work in sports and medical care and summarizes the existing datasets and models.
 \item We elaborate challenges and future opportunities of AQA to facilitate future research.
\end{itemize}

The rest of the survey is organized as follows. 
We first present the basic definition and form of AQA task in section \ref{ch2} and then summarize the problems and challenges faced in AQA in section \ref{ch3}.
Then we describe existing AQA datasets and metrics in section \ref{ch4}.
In section \ref{ch5}, we reviewed the current AQA work from two aspects: sports and medical care.
Finally, in section \ref{ch6}, we provide discussions and future research opportunities of AQA.

    \begin{figure}[tbp]
    \centerline{\includegraphics[width=\linewidth]{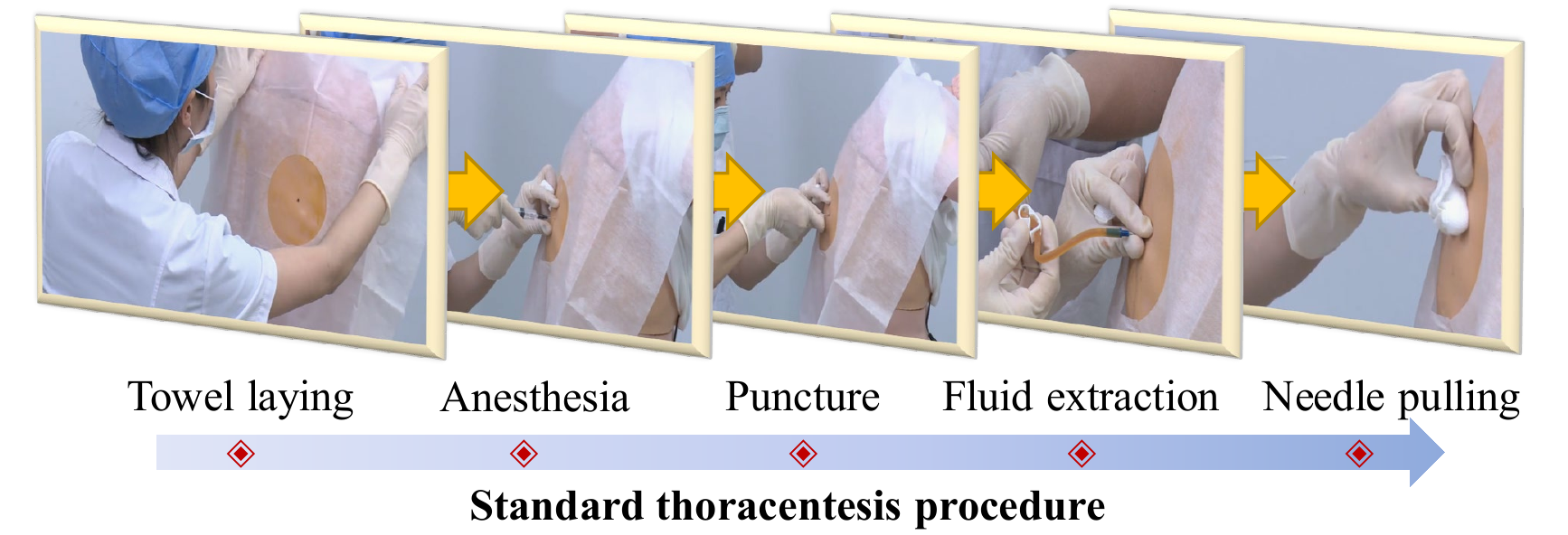}}
    \caption{Complexity of medical process: thoracentesis.}
    \label{fig1}
    \end{figure}

\section{Basic Form of AQA}\label{ch2}

Video-based AQA is a kind of action analysis task, which automatically completes the objective evaluation of specific human action quality through videos. 
The AQA model and human action recognition model have some similarities. They both complete the feature extraction first and then complete various specific tasks through the network head.
The general forms of AQA tasks could be classified into three types: regression scoring, grading and pairwise sorting.
For clarity, we will formalize the evaluation process as follow.

Given a set of video sequences $\mathbf{S}=\{ S_i \}, (i=1,2,\cdots,N)$ and each video can be represented as $S_i=\{ f_j \}, (j=1,2,\cdots,T)$, where the video set contains $N$ videos, video $S_i$ contains T frames and $f_j$ denotes the $j$-th frame in $S_i$.
The video will be sent to the feature extraction module and the action evaluation module, respectively.

In the \textit{feature extraction} phase, the traditional methods first complete the extraction and selection of spatio-temporal keypoints and then adopt Discrete Fourier Transform (DFT), Discrete Cosine Transform (DCT) or linear combination to complete feature aggregation.

However, deep learning methods usually adopt deep convolution networks (DCNs) and recurrent neural networks (RNNs) to complete video embedding. The feature extraction process can be denoted as: 
    \begin{equation}
    X=\mathcal{F}_{DL}(S_i), X\in\mathbb{R}^K
    \end{equation}
where $\mathcal{F}_{DL}(\cdot)$ denotes the deep neural network, such as I3D, C3D and other video feature extraction networks. $K$ represents the dimension of video features.

In the \textit{evaluation} phase, the obtained features are sent to the evaluation module of AQA. The evaluation module can be classified into the following three types:

\textbf{Regression scoring}. 
The form of regression scoring usually appears in sports.
Since the referees give the ground-truth scores of videos in AQA dataset in sports events, the dataset can be constructed through the broadcast videos of large-scale sports events, which is less challenging to obtain.
Support Vector Regression (SVR) model or Fully Connected Network (FCN) are adopted to complete score prediction directly. 
Mean Square Error (MSE) is usually used as the metric in regression scoring:
    \begin{equation}
        MSE=\frac{1}{N} \sum_{i=1}^{N}(\hat{s}_i-s_i)^2
    \end{equation}
where $\hat{s}_i$ denotes the prediction score of the AQA model on video $S_i$, and $s_i$ represents the ground-truth score of video $S_i$.

\textbf{Grading}. The form of grading usually appears in medical skill operation evaluation tasks. 
The operator's action will be divided into specific levels, such as \textit{novice}, \textit{medium} and \textit{expert}.
Therefore, the action quality assessment is transformed into a classification problem.
Given dataset $\mathbf{D}={\left(S_i,C_i\right)},\ (C_i\in{1,2,\cdots,M})$, where $C_i$ is the skill level label corresponding to video $S_i$ and there are $M$ skill levels in total. Classification accuracy is usually adopted as the metric.

\textbf{Pairwise sorting}. The pairwise sorting task takes any two videos from the video library to evaluate the action quality.
Suppose that the video dataset contains $N$ videos, so there are $C_{N}^{2}$ combinations in total.
The label matrix $M$ can be constructed according to the following rule:

    \begin{equation}
    M(i,j)=
        \left\{\begin{matrix}
        1,{\ \ \ Q(S}_i)>{Q(S}_j)\\ 
        -1,{\ Q(S}_i)<{Q(S}_j)\\ 
        0,\ {\ \ Q(S}_i)={Q(S}_j)
        \end{matrix}\right.
    \end{equation}  
where $Q(S)$ denotes the action quality assessment result of video $S$.
And the label matrix $M$ satisfies the following constraints:
    \begin{equation}
        M\left(i,j\right)=-M(j,i)
    \end{equation}
The model performance is measured by pairwise sorting accuracy.

    \begin{figure}[tbp]
    \centerline{\includegraphics[width=\linewidth]{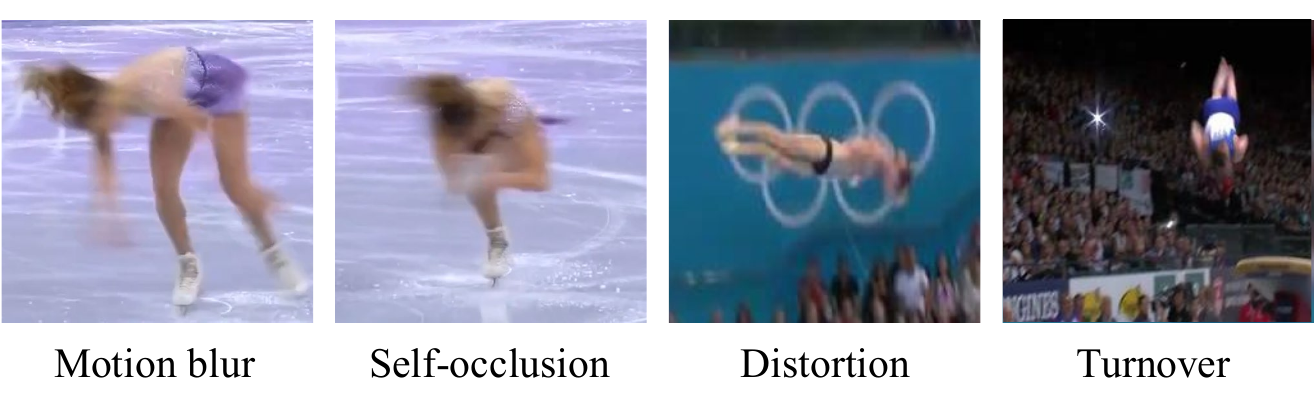}}
    \caption{The complexity of athletes' postures.}
    \label{fig2}
    \end{figure}

\section{Challenges}\label{ch3}
The challenges faced by AQA are highly related to its application background.
For example, the AQA system in sports focuses on the compliance of athletes' whole body and limb movements.
AQA system in medical care focuses on doctors' gestures, medical device interaction and task completions.
AQA system in industrial scenario focuses on the compliance and risk of workers' operation.
At the same time, all these AQA models also face some common challenges.
Therefore, the challenges faced by the AQA model can be classified into the following two categories:

            \begin{table*}
                \renewcommand\arraystretch{1.1}
              \caption{Summary of existing AQA datasets.}
              \label{tab1}
              \begin{tabular}{ccccccc}
                \toprule
                Scene & Dataset & Publishing & Action class & Size & Label form & Website \\
                
                \midrule
                \multirow{11}{*}{Sports} & MIT-Diving\cite{MIT-Diving}    & ECCV 2014     & 1     & 159   & AQA score & \href{https://www.csee.umbc.edu/~hpirsiav/quality.html}{https://www.csee.umbc.edu/~hpirsiav/quality.html}  \\
                & MIT-Skating\cite{MIT-Diving}   & ECCV 2014     & 1     & 159   & AQA score & \href{https://www.csee.umbc.edu/~hpirsiav/quality.html}{https://www.csee.umbc.edu/~hpirsiav/quality.html} \\
                & UNLV Dive\cite{UNLV-Dataset}     & CVPRW 2017    & 1     & 370   & AQA score & \href{http://rtis.oit.unlv.edu/datasets.html}{http://rtis.oit.unlv.edu/datasets.html} \\
                & UNLV Vault\cite{UNLV-Dataset}    & CVPRW 2017    & 1     & 176   & AQA score & \href{http://rtis.oit.unlv.edu/datasets.html}{http://rtis.oit.unlv.edu/datasets.html} \\
                & BPAD\cite{BPAD}          & ICCV 2017     & 1     & 48    & AQA score & ------ \\
                & AQA-7\cite{AQA-7}         & WACV 2018     & 7     & 1189  & AQA score & \href{http://rtis.oit.unlv.edu/datasets.html}{http://rtis.oit.unlv.edu/datasets.html} \\
                \cline{6-6}
                & MTL-AQA\cite{MTL-AQA}       & CVPR 2019     & 16    & 1412  & \makecell[c]{AQA score\\Action class\\Commentary} & \href{https://github.com/ParitoshParmar/MTL-AQA}{https://github.com/ParitoshParmar/MTL-AQA} \\
                \cline{6-6}
                & FisV-5\cite{FisV-5}        & TCSVT 2020    & 1     & 500   & TES, PCS & \href{https://github.com/loadder/MS\_LSTM.git}{https://github.com/loadder/MS\_LSTM.git} \\
                & FR-FS\cite{TSA-Net}         & ACM MM 2021   & 1     & 417   & Action class & \href{https://github.com/Shunli-Wang/TSA-Net}{https://github.com/Shunli-Wang/TSA-Net} \\
                \midrule
                
                Medical Care    & JIGSAWS \cite{JIGSAWS}   & MICCAI 2014     & 3   & 103 & AQA score & \href{http://cirl.lcsr.jhu.edu/jigsaws.}{http://cirl.lcsr.jhu.edu/jigsaws.} \\
                \midrule
                                
                \multirow{3}{*}{Daily Life} & EPIC-Skills\cite{Epic-skills} & CVPR 2018 & 3 & 196 & Pair-Rank & 
                \makecell[c]{\href{https://drive.google.com/file/d/1oX0dPM5IP638nB0YHt4L70aigIdqqpYr/}{https://drive.google.com/file/d}\\
                \href{https://drive.google.com/file/d/1oX0dPM5IP638nB0YHt4L70aigIdqqpYr/}{1oX0dPM5IP638nB0YHt4L70aigIdqqpYr/}} \\
                \cline{7-7}
                & BEST\cite{BEST}	& CVPR 2019	& 5	& 500	& Pair-Rank & 
                \href{https://github.com/hazeld/rank-aware-attention-network}{https://github.com/hazeld/rank-aware-attention-network} \\
                & Infant Grasp\cite{Infant-Grasp}	& arXiv 2019	& 1	& 94	& Pair-Rank & ------ \\
                
              \bottomrule
            \end{tabular}
            \end{table*}

\textbf{Challenges in specific areas}.
The task forms in different fields are quite different, and the problems they face also have their characteristics.
The action in medical care has the characteristics of high temporal complexity, semantic richness and low fault tolerance, which puts forward high requirements for the semantic understanding ability of the AQA system.
For example, the standard thoracentesis shown in Fig. \ref{fig1} consists of a series of sub-actions: towel laying, anesthesia, puncture, fluid extraction, needle pulling, etc.
However, AQA tasks in the field of sports mainly face problems such as body distortion and motion blur, as shown in Fig. \ref{fig2}.

\textbf{Common challenges}.
In addition, there are some common challenges in various fields, such as efficiency, view occlusion, model interpretability, etc.
In medical scenarios, the AQA system needs to give evaluation results and feedback in a real-time manner as much as possible to facilitate the tested personnel to correct unqualified actions in time.
In sports scenarios, self-occlusion during complex movements will seriously affect human body detection and pose estimation. And the same to medical operations.

There is still a long way to go between the existing models and those with high precision, high stability, strong anti-interference and excellent interpretability.
Therefore, the video-based action quality assessment system has a considerable exploration space and research significance.

\section{Datasets and Metrics}\label{ch4}
We review the video-based AQA datasets in sports and medical care and briefly sort out other datasets in daily life. Then we summarize the evaluation metrics commonly used in AQA tasks. All AQA datasets are listed in Tab. \ref{tab1}.

\subsection{AQA Datasets in Sports}
\textbf{MIT-Diving} \& \textbf{MIT-Skiing}\cite{MIT-Diving}. The MIT-Diving dataset contains 159 Olympic diving competition videos. The videos are played back in slow motion with a frame rate of 60fps. The whole dataset includes 25000 frames. Each video contains 150 frames on average, with a score range of 20 to 100. Researchers visited the MIT diving team to consult the adjustment direction of each body part for evaluating the feedback system.
The MIT-Skiing dataset contains 159 Olympic figure skating videos with a frame rate of 24fps. The whole dataset includes 630000 frames. Each video contains 4200 frames on average, with a score range of 0 to 100.

\textbf{UNLV Dive} \& \textbf{UNLV Vault}\cite{UNLV-Dataset}. The UNLV Dive dataset is expanded from MIT-Diving, which contains 370 videos. The UNLV Vault dataset contains 176 gymnastics videos, with an average length of about 75 frames and a score range of 0 to 20.

\textbf{Basketball Performance Assessment Dataset (BPAD)} \cite{BPAD}. The BPAD contains 48 videos in total. 
The first 24 videos were taken on the first day for training, and the last 24 videos were taken on the second day for testing. All videos are recorded from the first-person perspective.
Tag label information is obtained by pairwise sorting. After professional basketball players sort the videos in pairs, 250 pairs of training and 250 pairs of tests are finally generated.

\textbf{AQA-7}\cite{AQA-7}. The AQA-7 dataset contains seven sports and 1189 videos in total. The items and corresponding numbers are \textit{diving}-370, \textit{gymnastics}-176, \textit{snowing board}-175, \textit{skiing}-206, \textit{synchronous diving}-88 and \textit{trampoline}-83. The \textit{trampoline} category is usually excluded due to the long duration. Among all videos, 803 videos are used for training and 303 for testing.

\textbf{MTL-AQA} \cite{MTL-AQA}. The MTL-AQA dataset is currently the largest AQA dataset. 
This dataset 16 contains different events and 1412 diving samples, including single or synchronous diving, male or female athletes, 3m springboard or 10m platform.
In addition to providing ground-truth scores for each video, this dataset also provides fine-grained action classes and comment information, so it can be used to complete AQA, action recognition and comment generation.
The scores and difficulty degrees of all videos are given by 7 judges. Among all videos, 1059 videos are used for training and 353 for testing.

\textbf{FisV-5} \cite{FisV-5}. The FisV-5 dataset contains 500 figure skating competition videos, with an average duration of 2 minutes and 50 seconds. The score label of the video is given by 9 professional judges and is divided into two parts: Total Element Score (TES) and Total Program Component Score (PCS). Among all videos, 400 videos are used for training and 100 for testing.

\textbf{Fall Recognition in Figure Skating (FR-FS)}\cite{TSA-Net}. Existing AQA datasets of figure skating\cite{MIT-Diving, FisV-5} only contain long videos, which will lead to the inundation of detailed information, and finally affect the evaluation performance of the AQA model. 
To solve this problem, Wang \textit{et al.}\cite{TSA-Net} proposed FR-FS to recognize figure skating falls and planned to build a delicate granularity AQA system gradually. 
Videos in FR-FS contains the movements of the athlete's take-off, rotation, and landing. The FR-FS dataset contains 417 videos. Among these videos, 276 are smooth landing videos, and 141 are fall videos. 

\subsection{AQA Datasets in Medical Care}
Compared with sports, video-based AQA datasets are more scarce in medical care.
Each research institution builds its data collection platform to construct the AQA datasets, so there is very little open source-work. Currently, the only open-source work in medical care is JIGSAWS.

\textbf{JIGSAWS} \cite{JIGSAWS}. The full name of JIGSAWS is JHU-ISI Posture and Skill Assessment Working Set. JIGSAWS includes three surgical behaviors: \textit{suturing}, \textit{needle passing} and \textit{knot tying}.
Each video is scored in many aspects, such as operation and result quality. The final score is the summation of sub-scores.
All videos in the dataset are recorded with stereo cameras. Eight doctors with different skills are invited as operators. It contains 36 trials of knot tying, 28 trials of need passing and 39 trials of suturing.

\subsection{Other AQA Datasets}
In addition to the datasets in sports and medical care, some researchers have explored tasks in daily life and constructed AQA datasets. Since there is no strict operation process regulation for these actions, the evaluation metric of pair-wise ranking is adopted.

\textbf{Epic skills 2018}\cite{Epic-skills}. The Epic skills 2018 dataset consists of three sub-datasets: 
1. \textit{Dough-Rolling} is collected from the CMU-MMAC dataset \cite{CMU-MMAC} which contains actions performed 33 different participants.
2. \textit{Drawing}: Four participants were asked to complete painting imitations five times. Two reference images are provided: a cartoon of Sonic the Hedgehog and a gray scale photo of a hand.
3. \textit{Chopstick-Using}: Eight participants were asked to use chopsticks to carry the beans in the lunch box for five times. The action quality is determined by the successful transport times within one minute.

\textbf{BEST} \cite{BEST}. The full name of BEST is Bristol Everyday Skill Tasks. 
This dataset is constructed for skill evaluation of long-term video with an average duration of 188s.
It contains 500 videos in total and 5 daily tasks: \textit{scrambling eggs}, \textit{braiding hair}, \textit{tying a tie}, \textit{making an origami crane} and \textit{applying eyeliner}.

\textbf{Infinite grasp dataset}\cite{Infant-Grasp}. This dataset is introduced from psychological research and contains 94 videos of infants grabbing objects.
The length of the video varies from 80 to 500. 
Five psychologists were asked to label the videos in pairs, resulting in 4371 pairs, of which 3318 pairs with pairwise differences.

\subsection{Performance Metrics}
Due to the rich task forms, there is no general evaluation metric for AQA tasks.
In section 2, the AQA models are categorized into the following three types:
regression scoring, grading and pairwise sorting.
Therefore, the corresponding evaluation metrics are mean square error, classification accuracy and Spearman correlation coefficient $\rho$. The Spearman correlation coefficient is defined as follow.
\begin{equation}
    \rho = \frac{\sum_i(p_i-\bar{p})(q_i-\bar{q})}{\sqrt{\sum_i(p_i-\bar{p})^2}\sum_i(p_i-\bar{q})^2}
\end{equation}

            \begin{table*}
              \caption{Summary of existing AQA methods.}
              \label{tab1}
              \begin{tabular}{ccccccc}
                \toprule
                Ref. & Published & Dataset & Backbone & Network Head & Metrics & Website \\
                \midrule
                
                \cite{RO-MAN} & RO-MAN 2016 & Powerlifting & RSO Net & Feedback & Time Series Analysis & ------ \\ \cline{3-7}
                \cite{S3D} & ICIP 2018   & \makecell[c]{AQA-7\\JIGSAWS} & P3D & Reg & Spearman Rank Cor. & \href{https://github.com/YeTianJHU/diving-score}{https://github.com/YeTianJHU/diving-score} \\ \cline{3-7}
                \cite{e2e} & PCM 2018    & \makecell[c]{UNLV-Diving\\UNLV-Vault\\MIT-Skate} & C3D & Rank loss Reg & \makecell[c]{Spearman Rank Cor.\\Mean Euclidean Dis.} & ------ \\ \cline{3-7}
                \cite{AQA-7} & WACV 2019   & AQA-7 & C3D & Reg & Spearman Rank Cor. & \href{http://rtis.oit.unlv.edu/datasets.html}{http://rtis.oit.unlv.edu/datasets.html} \\ \cline{3-7}
                \cite{MTL-AQA} & CVPR 2019   & MTL-AQA & C3D & \makecell[c]{Caption\\Cls\\Reg} & Spearman Rank Cor. & \href{https://github.com/ParitoshParmar/MTL-AQA}{https://github.com/ParitoshParmar/MTL-AQA} \\ \cline{3-7}
                \cite{joint-relation-graphs} & ICCV 2019   & \makecell[c]{AQA-7\\JIGSAWS} & I3D & Reg & Spearman Rank Cor. & ------ \\ \cline{3-7}
                \cite{FisV-5} & TCSVT 2020  & \makecell[c]{FisV\\MIT-Skate} & C3D & Reg & \makecell[c]{Spearman Rank Cor.\\Mean Square Error} & \href{https://github.com/loadder/MS\_LSTM.git}{https://github.com/loadder/MS\_LSTM.git} \\ \cline{3-7}
                \cite{USDL} & CVPR 2020   & \makecell[c]{AQA-7\\MTL-AQA\\JIGSAWS} & I3D & Reg & Spearman Rank Cor. & \href{https://github.com/nzl-thu/MUSDL}{https://github.com/nzl-thu/MUSDL} \\ \cline{3-7}
                \cite{TSA-Net} & ACM MM 2021  & \makecell[c]{FR-FS\\AQA-7\\MTL-AQA} & I3D & \makecell[c]{Reg\\Cls} & Spearman Rank Cor. & \href{https://github.com/Shunli-Wang/TSA-Net}{https://github.com/Shunli-Wang/TSA-Net} \\

                \midrule
                \cite{BPAD} & ICCV 2017 & \makecell[c]{First-Person\\Basketball} & Att. to scale & Pair-Rank & Ranking Acc. & ------ \\\cline{3-7}
                \cite{Epic-skills} & CVPR 2018 & \makecell[c]{EPIC-Skills\\JIGSAWS} & TSN & Pair-Rank & Ranking Acc. & ------ \\\cline{3-7}
                \cite{BEST} & CVPR 2019 & \makecell[c]{BEST\\EPIC-Skills} & I3D & Pair-Rank & Ranking Acc. & \makecell[c]{\href{https://github.com/hazeld/rank-aware-attention-network}{https://github.com/hazeld/}\\              \href{https://github.com/hazeld/rank-aware-attention-network}{rank-aware-attention-network}} \\\cline{3-7}
                \cite{Infant-Grasp} & ICCVW 2019 & \makecell[c]{Infant Grasp\\EPIC-Skills\\ JIGSAWS} & ResNet101 & Pair-Rank & Ranking Acc. & ------ \\
                
              \bottomrule
            \end{tabular}
            \end{table*}

\section{AQA Models}\label{ch5}

    \begin{figure}[tbp]
    \centerline{\includegraphics[width=\linewidth]{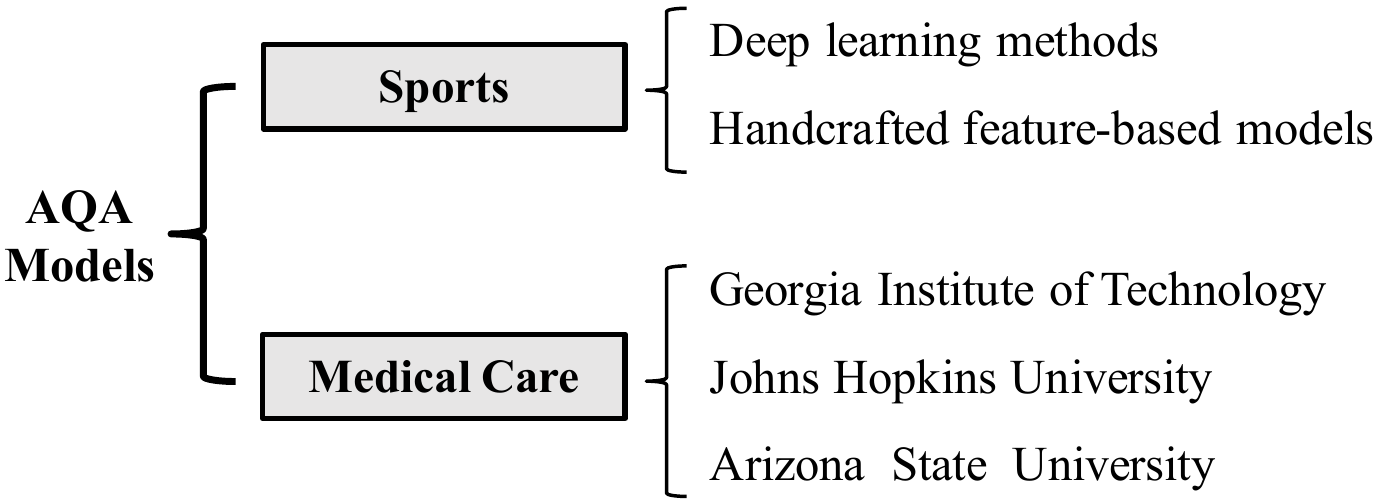}}
    \caption{Categorization methodology adopted by this paper.}
    \label{fig3}
    \end{figure}

From the perspective of development history, the emergence of AQA models in medical care is earlier than in sports because medical skills have two characteristics. 
On the one hand, experienced doctors have to evaluate the operation skills of each novice in the standard resident training process, and the workload is enormous.
On the other hand, with the continuous development of medical robots (such as \textit{Da Vinci Robot}), robots can replace doctors to complete some medical operations to a certain extent. It is necessary to specify a recognized standard to evaluate robot operations.
Before the prosperity of deep learning, these factors have aroused people's interest in skills assessment in medical care.
Generally speaking, most medical skill evaluation methods use traditional features to model video information. We will describe these efforts in detail below.

Compared with medical care, researches on action quality assessment on sports video started relatively late. 
These studies mainly focus on the actions of athletes in the Olympic Games, such as diving, gymnastics, trampoline and other sports mentioned above.
The rise of deep learning in recent years provides a powerful tool for sports video assessment. 
Most studies have achieved considerable results through CNN and RNN. This is also the development trend of future research.
Considering the above internal differences between sports and medical care, we still take the research topic of the paper as a taxonomic basis. 
Fig. \ref{fig3} shows the categorization methodology adopted by this paper.

\subsection{AQA Models in Sports}
In this subsection, we focus on the action quality assessment methods in sports video. We divide these methods into two categories: models based on deep learning and handcrafted features. 
Note that since the models tested on the daily life datasets\cite{Epic-skills, BEST, Infant-Grasp} are all based on deep learning, we divide them into sports, even if these methods are not oriented to professional sports.

\subsubsection{Models based on deep learning} 
\begin{figure}[tbp]
    \centerline{\includegraphics[width=\linewidth]{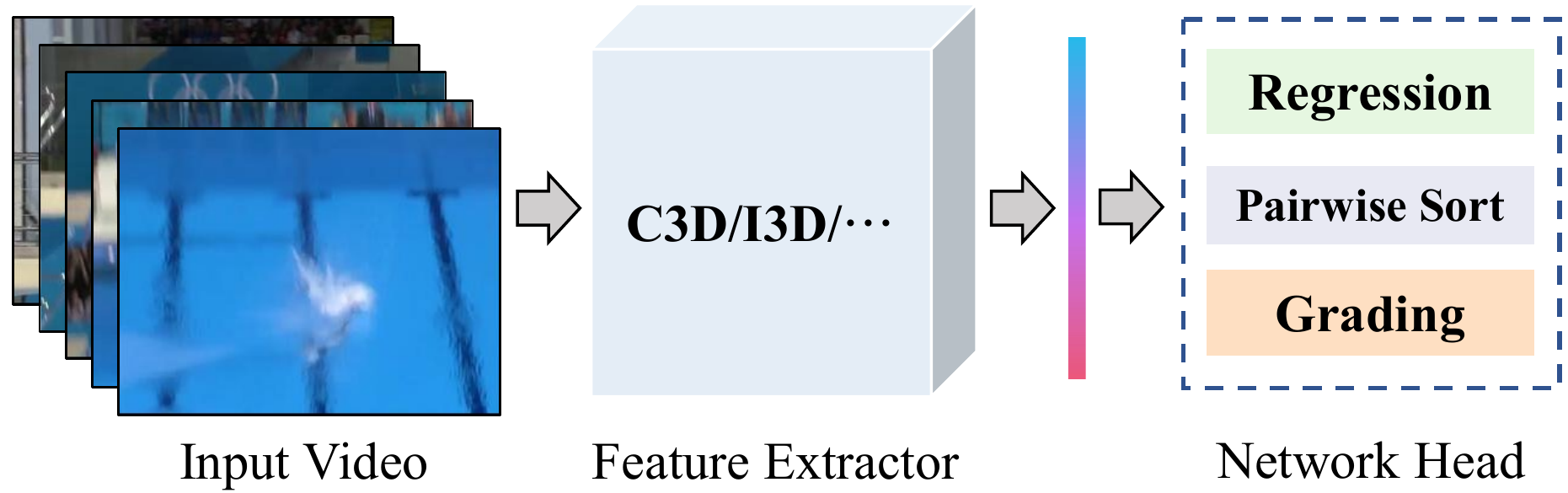}}
    \caption{Framework of AQA model based on deep learning.}
    \label{fig4}
    \end{figure}

Deep learning methods usually adopt 2D-CNN, 3D-CNN and LSTM to extract and aggregate the features of videos and then complete the assessment task.
As shown in Fig.\ref{fig4}, these deep learning-based AQA frameworks are composed of a video feature extraction module and an evaluation module. The form of the evaluation module is highly related to the type of assessment task.
According to the framework position concerned by these works, these methods can be categorised into the following two classes: \textit{structure designing} and \textit{loss designing}.

Some studies focus on the design of the network structure to extract more discriminative features.
In order to learn the local and global sequential information in long-term videos, Xu \textit{et al.} \cite{FisV-5} proposed Self-Attentive LSTM and Multi-scale Convolutional Skip LSTM to predict TES and PCS in figure skating. 
Xiang \textit{et al.} \cite{S3D} divided the diving process into four stages:
beginning, jumping, dropping and entering into the water. They adopted four independent P3D\cite{P3D} models to complete feature extraction.
Pan \textit{et al.} \cite{joint-relation-graphs} proposed a graph-based joint relation model to analyze the motion of human nodes. Human nodes and their correlation are modelled by the proposed Joint Commonality Module and the Joint Difference Module.
Parisi \textit{et al.} \cite{RO-MAN} proposed a new recurrent neural network, which adopts the growing self-organizing structure to learn the body motion sequence and complete the matching.
Kim \textit{et al.} \cite{EvaluationNet} modeled the action as a structured process. Action units are encoded from dense trajectories with a LSTM network.
Wang \textit{et al.} \cite{TSA-Net} proposed a feature aggregation mechanism named tube self-attention module to efficiently generate rich spatio-temporal contextual information by adopting sparse feature interactions.

Some studies focus on the design of network loss to obtain better action quality assessment performance.
Li \textit{et al.} \cite{e2e} proposed an end-to-end framework that takes C3D as the feature extractor and designs a ranking loss integrated with the MSE loss.
Parmar \textit{et al.} \cite{MTL-AQA} explored the AQA model in the multi-task learning scenario. They proposed three parallel prediction tasks: action recognition, comment generation, and AQA score regression, and then verified the effectiveness by the proposed C3D-AVG-MTL framework.
Tang \textit{et al.} \cite{USDL} proposed an uncertainty-aware score distribution learning approach. They incorporated the difficulty degree into the modeling process and simulated the scoring process more realistically.

In addition, some studies focus on the quality comparison task of paired actions. 
Bertasius \textit{et al.} \cite{BPAD} proposed a model based on the first perspective videos in basketball games. This model adopted a convolutional-LSTM network to detect events and evaluate the quality of any two movements.
Doughty \textit{et al.} \cite{Epic-skills, BEST} proposed the task of sorting paired videos. These two works proposed EPIC-Skills and BEST,  respectively. 
In \cite{Epic-skills}, they took temporal segment network (TSN)\cite{TSN} as the backbone and proposed a ranking loss based on margin loss.
\cite{BEST} is the subsequent work of \cite{Epic-skills}, which released the BEST dataset with richer scene information and proposed a model with a more complex sorting loss function based on I3D\cite{I3D}.
Li \textit{et al.} \cite{Infant-Grasp} proposed a spatial attention model based on RNN. Attention pooling and temporal aggregation mechanisms were adopted to complete feature extraction. This model was tested on the released Infant Grasp dataset.

\subsubsection{Models based on handcrafted features}
Traditional AQA methods take handcrafted features as discriminant features. These feature extraction methods mainly appeared before 2014.
In \cite{AI-ED1995}, Gordon first proposed video-based AQA tasks and took the gymnastics scoring as an example to analyse the realizability of AQA technology.
Ilg \textit{et al.} \cite{STCorr2003} established correspondence at the global action sequence level and the single motion element level. A spatio-temporal deformation model is proposed for example comparison. 
Çeliktutan \textit{et al.} \cite{ACM2013} proposed a dynamic skeleton sequence alignment method based on graph and applied it to action recognition and assessment.
Pirsiavash \textit{et al.} \cite{MIT-Diving} proposed an analysis method in the frequency domain to complete the quality evaluation based on the combination of low-level visual features and high-level pose features.
Paiement \textit{et al.} \cite{BMVC2014} proposed a general method for on line assessment of motion quality based on Kinect data. After reducing the dimension of noisy data by robust nonlinear manifold learning, the matching degree between observation data and prediction data is calculated frame by frame based on the Markov hypothesis.

\subsection{AQA Models in Medical Care}
Some institutions have made a series of contributions to medical skills assessment, such as Johns Hopkins University (JHU), Georgia Institute of Technology (GIT), Arizona State University (ASU), etc. 
Interestingly, compared with the action quality assessment of sports video, it is easy to find that these works have a strong inheritance. 
Therefore, we will creatively review the literature according to the publishing organization as the classification basis in this part.

\begin{figure}[tbp]
    \centerline{\includegraphics[width=\linewidth]{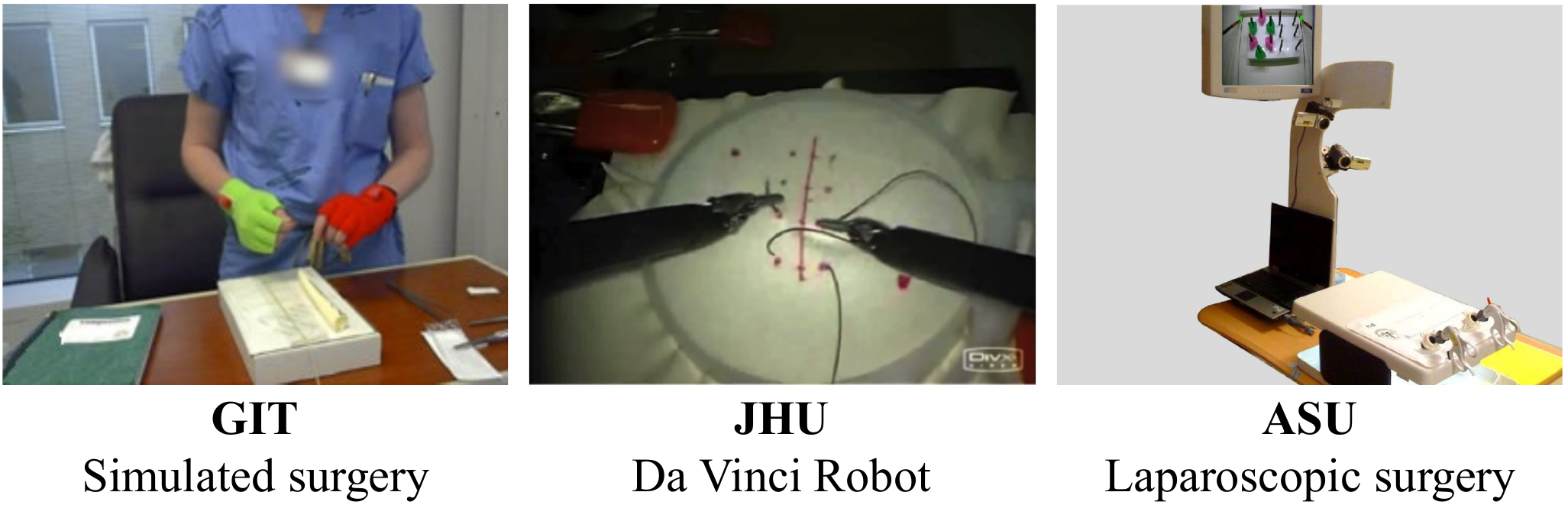}}
    \caption{Dataset acquisition platforms of three three research series.}
    \label{fig5}
    \end{figure}

Due to the strong professionalism of medical skill assessment, there is no unified dataset as a benchmark.
The series of studies of the institutions mentioned above only focus on a specific scenario. 
Studies of GIT focus on the medical behavior quality assessment under the Objective Structured Assessment of Technical Skill (OSATS) system\cite{OSATS}. Studies of GIT focus on the action quality assessment of the surgical robots. Studies of ASU focus on the action quality assessment using surgical training equipment.
Fig. \ref{fig5} shows the scenarios concerned by these three series of studies.
Next, we will introduce these three types of studies in chronological order respectively.

\subsubsection{GIT}
The series studies of Georgia Institute of Technology focus on the Surgical skills evaluation under the OSATS system\cite{OSATS}.
Inspired by LDS and bag-of-features (BoF) methods \cite{MICCAI2016, 2013Surgical-27}, Sharma \textit{et al.} \cite{M2CAI2016} proposed an enhanced BoW model to solve the problem that traditional models cannot process implicit temporal information and causality.
In\cite{M2CAI2016, ISBI2014}, Sharma \textit{et al.} evaluated medical skills through video motion texture and sequential motion texture modeling respectively.
Based on\cite{M2CAI2016}, Sharma \textit{et al.}\cite{ISBI2014} introduced a data-driven temporal window to complete the modeling of sequential motion features.
Zia \textit{et al.}\cite{MICCAI-2015,IJCARS-2016} adopted the same methods, and the latter is the extended journal version of the former. 
In \cite{MICCAI-2015}, after the extraction of STIP keypoints, K-means clustering is adopted to generate time series, and then DCT and DFT are utilized to explore the temporal characteristics.
To solve the problem of deficient information in traditional methods, Zia \textit{et al.}\cite{IJCARS-2017} proposed an off-line unsupervised temporal clustering method and divided the long-term surgical operation into small steps, which is similar to those studies of JHU.
Zia \textit{et al.}\cite{IJCARS-2018} created a surgical skill evaluation dataset based on camera and accelerometer and proposed a model that combines approximate entropy and cross approximate entropy to quantify the predictability and regular beat of time series.

\subsubsection{JHU}
The series of studies of Johns Hopkins University focused on the scene of robotic minimally invasive surgery (RMIS). 
Most of the studies take \textit{Da Vinci Robot} as the equipment for data acquisition.
Haro \textit{et al.} \cite{MICCAI2016} and Zappella \textit{et al.} \cite{2013Surgical-27} explored the automatic classification of gestures and skills in robot assisted surgery.
Haro \textit{et al.} \cite{MICCAI2016} proposed a surgical posture classification method based on linear dynamic system (LDS) and bag-of-features and carried out more experiments in the extended journal version \cite{2013Surgical-27}. Although these two studies did not directly assess the operation quality, they laid a solid foundation for follow-up work.
Malpani \textit{et al.} \cite{Pairwise2014} proposed a model that automatically scores surgical segments and completes the quality ranking by pairwise comparison.
Gao \textit{et al.} \cite{JIGSAWS} constructed and released the JIGSAWS dataset and expounded the concept of \textit{JHU Language of Surgery} project in detail.
Some early studies complete the evaluation of medical skills only based on kinematic data. For example, robot operation assessment based on HMM\cite{SparseHMM}, temporal modeling of kinematic data based on RNN\cite{RNN-MICCAI2016}, endoscopic sinus surgery pathway evaluation \cite{Minimally-Invasive2012}, and skill evaluation in septoplasty\cite{septoplasty2015}.

\subsubsection{ASU}
The series of studies of Arizona State University mainly focus on the fundamentals of laparoscopic surgery (FLS-Box).
Traditional medical skill evaluation methods usually need contact sensors which will affect the operation of doctors. 
To overcome this disadvantage, \cite{Islam2011, IslamIUI2013} proposed a surgical skill method based on visual information.
Based on Islam \textit{et al.} \cite{Islam2011,IslamIUI2013} improved the algorithm to real-time, added feedback function and tested the model in the real surgical video.
In order to deal with the semantic gap between low-level visual features and high-level surgical skills, Chen \textit{et al.} \cite{MMVR2013} fused multi-modal information and completed meaningful surgical skill analysis by looking for hidden space.
Zhang \textit{et al.} \cite{ZQ1, ZQ2, ZQ3} adopted HMM and its variants to solve AQA problems and made a series of progress.
In \cite{ZQ1}, they proposed a hierarchical Dirichlet process Markov model for the classification of surgical skills.
In \cite{ZQ2}, they proposed a relative HMM model, which only takes the relative ranking between paired videos as input for training.
Based on \cite{ZQ2}, researchers in \cite{ZQ3} improved the model and extended the application scenario to a language-based emotion recognition task.
Based on the previous work, Chen \textit{et al.} \cite{ICME2015} further integrated and improved the function of the action quality assessment system and improved the interactivity.

\subsection{Summary of AQA models}
We have reviewed the work related to AQA in sports and medical care. 
Through the above classification and summary, we can get the following findings.
\begin{itemize}
    \item From the perspective of development history, the AQA problem in medical care was put forward earlier than that in sports. 
    This is because the system that can automatically and objectively evaluate surgical skills has more urgent needs and greater practical significance.
    
    \item From the perspective of feature extractor, most of the methods before 2014 constructed the AQA system through handcrafted features and traditional classifiers.
    In recent years, some general network backbones have been proposed \cite{FineGym,epic-kitchens,LSTA,2019Forecasting}. More and more work began to use DNNs to build AQA model due to the stronger representation ability and achieved excellent performance.
    
    \item From the perspective of the datasets, it is more challenging to construct datasets in medical care than in sports. Datasets in sports usually consist of Olympic Games, while datasets in medical care are more professional and have more institutional background and continuity.
\end{itemize}

    \begin{figure}[tbp]
    \centerline{\includegraphics[width=\linewidth]{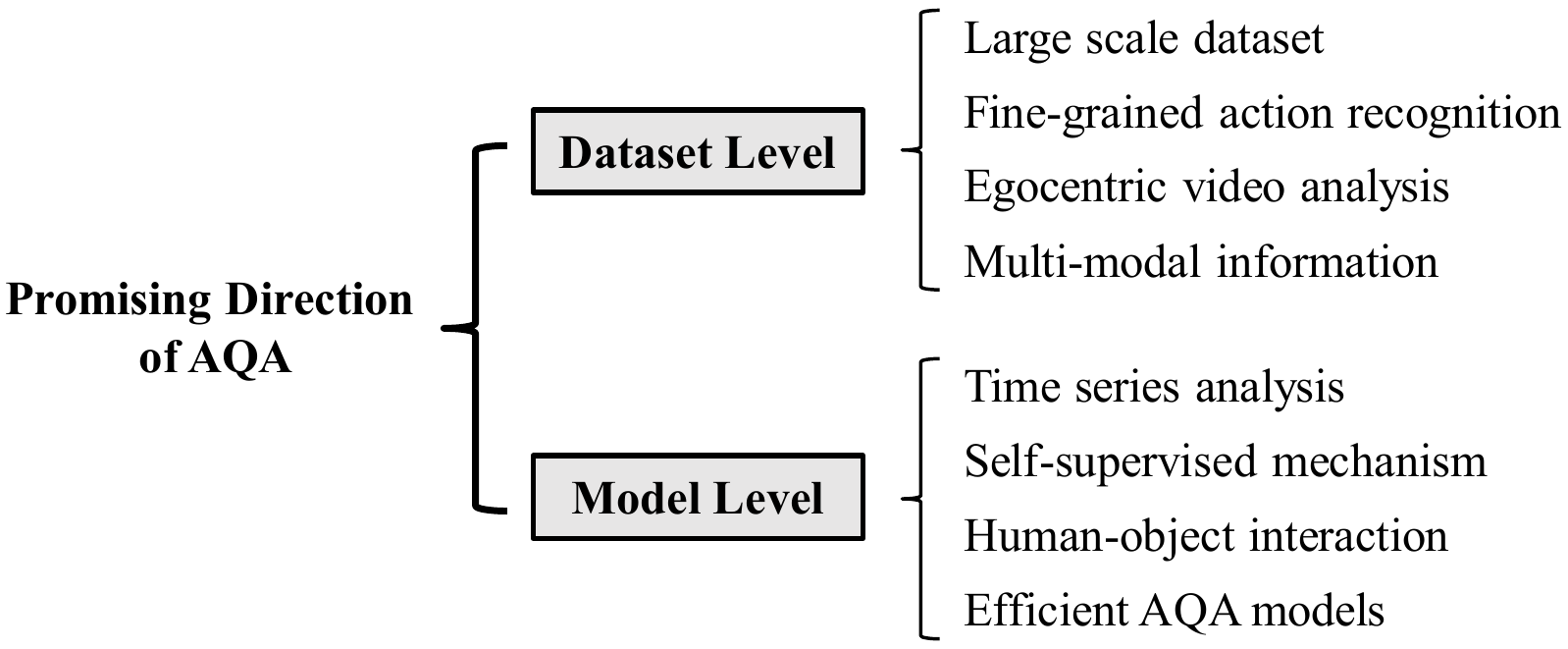}}
    \caption{Some promising research directions of AQA task in future.}
    \label{fig6}
    \end{figure}

\section{Prospect of Future Work}\label{ch6}
Combined with recent progress in human action understanding, this section looks forward to the future research direction of AQA task from two aspects: dataset and model, as shown in Fig.\ref{fig6}.
At the dataset level, existing AQA datasets are small in scale and have weak semantic richness.
We look forward to the proposal of datasets with larger scale and richer semantic information in the future. 
In addition, some novel datasets and models have been proposed recently in HAR tasks. 
These studies focus on fine-grained action recognition \cite{FineGym, epic-kitchens}, egocentric action recognition \cite{LSTA, 2019Forecasting} and multimodal information \cite{Unsupervised2020, 2019Self}.
New scenarios and datasets can promote the exploration and development of AQA.

At the model level, there are still many efforts to be paid for efficient and accurate AQA models. 
For example, how to make full use of temporal information to model the action, use unsupervised methods to avoid expert annotation, take human-object interaction into account, and design efficient models.
Recent studies in HAR has made a preliminary exploration on these problems, such as effective temporal modeling \cite{2018Temporal, SlowFast}, self-supervised mechanism\cite{SpeedNet, Self-Supervised2020}, taking human-object interaction information into consideration\cite{Something-Something, Egocentric}, and efficient designing of HAR models\cite{X3D, TSM}. 
These methods provide a solid foundation for the follow-up development of AQA.

Most existing AQA methods assume that the operation sequence is known and the task is relatively easy. There is no work to explore the long-term complex action quality assessment.
However, some studies have explored unsupervised structural modelling of complex actions, such as instruction video analysis \cite{instruction1, instruction2} and unsupervised process self discovery\cite{unsupervised-CVIU}.
These works explore the task of video analysis under more stringent conditions and put forward higher and requirements for AQA models.

\section{Conclusion}
Video-based action quality assessment is a critical and challenging problem in computer vision and has received considerable attention.
As the first comprehensive survey of AQA tasks, this paper summarizes the AQA datasets and methods according to the categorization methodology of sports and medical care. 
The challenges of the AQA task in two fields are discussed in detail.
In the end, we discuss the future development direction combined with recent studies.
Despite the considerable performance, existing AQA methods can only analyse the video under ideal conditions and can not handle complex scenes.
We believe that high-performance AQA systems will bring great convenience to our life, especially in sports, medical care and skill training.


\clearpage
\bibliographystyle{IEEEtran}
\bibliography{mylib.bib}
\vspace{12pt}

\end{document}